\documentclass[11pt,a4paper]{article}
\usepackage[utf8]{inputenc}
\usepackage[T1]{fontenc}
\usepackage{amsmath,amssymb,amsthm}
\usepackage{graphicx}
\usepackage{hyperref}
\usepackage{natbib}
\usepackage[margin=1in]{geometry}
\usepackage{booktabs}
\usepackage{caption}
\usepackage{subcaption}

\newtheorem{proposition}{Proposition}
\newtheorem{definition}{Definition}
\newtheorem{remark}{Remark}

\title{\textbf{The Novelty Bottleneck:} \\ A Framework for Understanding Human Effort Scaling in AI-Assisted Work}
\author{Jacky Liang}
\date{March 2026}

\begin{document}
\maketitle

\begin{abstract}
We propose a stylized model of human-AI collaboration that isolates a mechanism we call the \emph{novelty bottleneck}: the fraction of a task requiring human judgment creates an irreducible serial component analogous to Amdahl's Law in parallel computing. The model assumes that tasks decompose into atomic decisions, a fraction $\nu$ of which are ``novel'' (not covered by the agent's prior), and that specification, verification, and error correction each scale with task size. From these assumptions, we derive several non-obvious consequences: (1) there is no smooth sublinear regime for human effort---it transitions sharply from $O(E)$ to $O(1)$ with no intermediate scaling class; (2) better agents improve the coefficient on human effort but not the exponent; (3) for organizations of $n$ humans with AI agents, optimal team size \emph{decreases} with agent capability; (4) wall-clock time achieves $O(\sqrt{E})$ through team parallelism but total human effort remains $O(E)$; and (5) the resulting AI safety profile is asymmetric---AI is bottlenecked on frontier research but unbottlenecked on exploiting existing knowledge. We show these predictions are consistent with empirical observations from AI coding benchmarks, scientific productivity data, and practitioner reports. Our contribution is not a proof that human effort must scale linearly, but a framework that identifies the novelty fraction as the key parameter governing AI-assisted productivity, and derives consequences that clarify---rather than refute---prevalent narratives about intelligence explosions and the ``country of geniuses in a data center.''
\end{abstract}

\section{Introduction}

Aschenbrenner's ``Situational Awareness'' envisions hundreds of millions of AI agents compressing a decade of research into a year \citep{aschenbrenner2024}. Amodei's ``Machines of Loving Grace'' describes a ``country of geniuses in a data center'' compressing a century of biological progress into a decade \citep{amodei2024}. These visions raise a natural question: as AI agents grow more capable, how does the human effort required to direct them scale with task complexity?

One might expect human effort to become a vanishing fraction of total work. A 10$\times$ more capable agent should require less than $1/10$th the human involvement of its predecessor, yielding sublinear or even logarithmic scaling. Yet a recurring experience among practitioners---particularly in AI-assisted coding, but likely more broadly---tells a different story. The human time spent specifying intent, verifying outputs, and correcting errors remains stubbornly proportional to the size of the task. Agents make you faster, but twice the task still means roughly twice the human effort.

Why? And is this a temporary limitation of current tools, or something more structural?

In this paper, we construct a simple model to investigate this question. We do not claim to prove a law of nature. Like Amdahl's Law in parallel computing \citep{amdahl1967} --- which is ultimately just algebra on the definition of ``serial fraction'' but which reframed decades of processor design decisions --- our model's value lies not in its formal depth but in the \emph{mechanism it isolates} and the \emph{non-obvious consequences it derives}.

The model (Section~\ref{sec:model}) decomposes human effort into specification, verification, correction, and decomposition, each parameterized by a \emph{novelty fraction} $\nu$: the share of decisions not covered by the agent's prior knowledge. We make five explicit assumptions, discuss what happens when each breaks (Section~\ref{sec:assumptions}), and derive several consequences that are not obvious from the assumptions alone:

\begin{enumerate}
\item Under the model, there is no smooth sublinear regime for human effort. $H$ transitions sharply from $O(E)$ to $O(1)$, with nothing in between (Section~\ref{sec:model}).
\item Better agents improve the coefficient on $H$ but not the scaling exponent---a constants-vs-asymptotics distinction with practical implications (Section~\ref{sec:simulations}).
\item In organizations of $n$ humans with agents, optimal team size \emph{decreases} as agents improve, because amplified throughput increases coordination overhead. Wall-clock time achieves $O(\sqrt{E})$ but total human effort remains $O(E)$ (Section~\ref{sec:org}).
\item AI agents amplify \emph{exploitation} of existing knowledge but do not accelerate \emph{exploration} at the frontier, creating an asymmetric AI safety profile (Section~\ref{sec:implications}).
\end{enumerate}

Importantly, this framework does not refute the vision of AI-accelerated progress. Amodei's estimate of 5--10 years to compress a century of biological progress is \emph{consistent} with our model: it corresponds to a large constant-factor improvement on the routine fraction of scientific labor. What the framework clarifies is why the estimate is 5--10 years rather than 5--10 months---the serial insight chain that connects routine phases cannot be compressed by parallelism---and what organizational structures best exploit the acceleration that \emph{is} available.

\section{Model and Assumptions}
\label{sec:model}

We present a stylized model in the tradition of Amdahl's Law \citep{amdahl1967} and Acemoglu's task-based framework \citep{acemoglu2024}: deliberately simplified to isolate a specific mechanism, with assumptions stated explicitly so the reader can judge which conclusions survive relaxation.

\subsection{Definitions}

We model a task $T$ as a sequence of $E$ atomic decision units $d_1, d_2, \ldots, d_E$. Each decision requires some intent --- a specification of what the correct outcome is. The total intent of the task is $I(T)$, measured in bits.

\begin{definition}[Human Effort]
Human effort $H$ is the sum of four components:
\begin{equation}
H = H_{\text{spec}} + H_{\text{verify}} + H_{\text{correct}} + H_{\text{decompose}},
\end{equation}
where $H_{\text{spec}}$ is specification effort (communicating intent), $H_{\text{verify}}$ is verification effort (checking outputs), $H_{\text{correct}}$ is correction effort (fixing errors), and $H_{\text{decompose}}$ is decomposition effort (breaking the task into agent-sized pieces).
\end{definition}

\begin{definition}[Mutual Information / Shared Prior]
Let $P_{\text{agent}}$ be the agent's prior distribution over correct decisions and $P_{\text{human}}$ be the human's actual intent. The mutual information $M = I(P_{\text{agent}}; P_{\text{human}})$ captures how much of the human's intent the agent can infer without explicit communication.
\end{definition}

\begin{definition}[Novelty]
The novelty parameter $\nu \in [0,1]$ is the fraction of decisions where the agent's prior has high entropy --- where the agent cannot reliably predict the human's intent. A task is ``routine'' when $\nu \approx 0$ and ``novel'' when $\nu \approx 1$.
\end{definition}

\subsection{The Specification Bottleneck}

The total intent of the task satisfies $I(T) \sim O(E)$: each decision unit contributes some bits of intent. The human must explicitly provide only those bits not inferrable from the shared prior:
\begin{equation}
H_{\text{spec}} = I(T) - M \approx \nu \cdot E.
\end{equation}
For $H_{\text{spec}}$ to be sublinear, we need $M \geq I(T) - o(E)$, meaning the agent must be able to infer almost all intent bits. This is possible when the task is routine ($\nu \to 0$), but by definition impossible when the task is novel ($\nu$ bounded away from zero).

\subsection{The Verification Bound}

Even if specification were free, the human must verify agent output. The output has size $O(E)$, and any meaningful verification requires inspecting at least a fraction of it:
\begin{equation}
H_{\text{verify}} \geq c_v \cdot E,
\end{equation}
for some constant $c_v > 0$ that depends on the human's trust in the agent. A perfectly trusted agent (one that self-verifies reliably) would drive $c_v \to 0$, but calibrating that trust itself requires $O(E)$ experience.

\subsection{The Trajectory Divergence Model}

We model the agent's execution as a trajectory $x_t$ that deviates from the intended trajectory $x^*_t$ by accumulated errors:
\begin{equation}
x_t = x^*_t + \sum_{i=1}^{t} \epsilon_i, \quad \epsilon_i \sim \mathcal{N}(0, \sigma^2).
\end{equation}
The expected maximum deviation after $E$ steps is $O(\sigma\sqrt{E})$. To keep deviation bounded by tolerance $\delta$, the human must intervene (checkpoint) every $O(\delta^2/\sigma^2)$ steps, giving a total number of checkpoints proportional to $E$. This yields $H_{\text{correct}} \sim O(E)$ for any fixed error rate and tolerance.

\subsection{Assumptions and Boundary Conditions}
\label{sec:assumptions}

The model above rests on five assumptions. We state them explicitly because the conditions under which our thesis could be falsified correspond precisely to the conditions under which these assumptions break.

\textbf{A1: Decision independence.} We model the task as $E$ approximately independent decision units, each contributing $O(1)$ bits of intent. \emph{If this breaks}: hierarchical tasks allow a single high-level specification to resolve many downstream decisions. This compresses $H_{\text{spec}}$ by a constant factor (potentially large), but the hierarchy has $O(\log E)$ levels with branching, so total specification remains at least $\Omega(E^{1-\epsilon})$ in practice.

\textbf{A2: Binary novelty.} We treat each decision as either ``novel'' (requiring human input) or ``routine'' (agent handles it), with a constant fraction $\nu$ being novel. \emph{If this breaks}: in reality, novelty is a spectrum. Some decisions are partially inferrable, and an agent might get them right 70\% of the time rather than 0\% or 100\%. This would soften the sharp $O(E)$-to-$O(1)$ transition but would not change the asymptotic scaling class, because any positive probability of requiring human input per decision still produces a linear term.

\textbf{A3: Verification requires inspection.} We assume verification effort scales with output size. \emph{If this breaks}: formal verification, automated testing, or other machine-checkable criteria could reduce $H_{\text{verify}}$ to $O(1)$. Tasks with clear correctness criteria (code with tests, math with proofs) are rapidly automatable; tasks requiring human judgment (strategy, design, writing) resist automation regardless of agent capability. We formalize this as the ``verifiability dimension'' in Section~\ref{sec:verifiability}.

\textbf{A4: Human intent is exogenous.} We assume the human knows what they want and communicates it to the agent. \emph{If this breaks in one direction}: for exploratory work, intent is generated \emph{during} execution. This actually \emph{strengthens} our thesis---the human must be in the loop not just to specify but to react, making $H$ at least $O(E)$. \emph{If this breaks in the other direction}: an agent that develops its own intent and pursues its own goals doesn't need human specification at all. This is the autonomous AI scenario, which is outside our model's scope.

\textbf{A5: No within-task learning.} We assume the agent's prior is approximately fixed during a single task execution. \emph{If this breaks}: an agent that learns during execution could reduce effective $\nu$ dynamically. If $\nu(t) \sim 1/t$, then $H_{\text{spec}} \sim O(\log E)$---genuinely sublinear. This is the ``continual learning'' scenario and is the most likely route to eventual falsification of our framework. Current agents update their context window but do not update their weights or generalization capability mid-task.

\subsection{Direct Consequence: Linear Human Effort}

Given assumptions A1--A5, the linearity of $H$ follows immediately:
\begin{equation}
H = \nu \cdot E + c_v \cdot E + c_c \cdot E + c_d \cdot E = (\nu + c_v + c_c + c_d) \cdot E \sim O(E),
\end{equation}
where $c_v, c_c, c_d \geq 0$ are constants depending on agent capability. This is essentially bookkeeping---four $O(E)$ terms sum to $O(E)$. The linearity is \emph{assumed into the model} through A1 (each decision costs $O(1)$) and A2 (a constant fraction $\nu$ requires human input).

What is \emph{not} assumed but follows as a derived consequence is the following:

\begin{proposition}[No Sublinear Regime]
Under assumptions A1--A5, $H$ transitions sharply from $O(E)$ to $O(1)$ with no intermediate sublinear scaling class. Specifically, for $H$ to be $o(E)$, \emph{all} of the following must hold simultaneously: (i) $\nu = 0$, (ii) $c_v = 0$, (iii) $c_c = 0$, and (iv) $c_d = 0$.
\end{proposition}

This is non-trivial because one might expect that partially reducing $\nu$ or improving one of the $c$ terms would yield an intermediate regime (say, $H \sim O(\sqrt{E})$). It does not. The model predicts a binary outcome: either the agent can fully handle the task ($H = O(1)$), or it cannot and $H$ remains linear. Partial improvement only changes the coefficient.

\begin{remark}
The absence of an intermediate regime is the model's sharpest prediction and the one most likely to be wrong. If assumption A1 breaks (hierarchical task structure), there could be an $O(E^{1-\epsilon})$ regime. If A5 breaks (within-task learning), there could be an $O(\log E)$ regime. We discuss these possibilities in Section~\ref{sec:assumptions}.
\end{remark}

\subsection{The Verifiability Dimension}
\label{sec:verifiability}

Novelty alone is insufficient to characterize the human effort bottleneck. Following Karpathy \citep{karpathy2025verify}, we introduce a second dimension: \emph{verifiability} $v \in [0,1]$, the degree to which output correctness can be checked automatically (this extends assumption A3). This creates a $2 \times 2$ taxonomy:

\begin{itemize}
\item \textbf{Low $\nu$, high $v$} (routine bug fixes, refactoring): Agent handles autonomously. $H \approx O(1)$.
\item \textbf{High $\nu$, high $v$} (novel algorithm with test suite): Agent can iterate via self-correction. $H \sim \nu' E$ where $\nu' = \nu(1-\alpha v)$ for self-correction rate $\alpha$. Linear but with reduced coefficient.
\item \textbf{Low $\nu$, low $v$} (boilerplate writing, standard reports): Agent produces output but human must review. $H \sim c_v E$.
\item \textbf{High $\nu$, low $v$} (frontier research, product strategy): Agent cannot self-correct and human must both specify and verify. $H \sim (\nu + c_v)E$. This is the hardest quadrant.
\end{itemize}

Verifiability modulates the \emph{coefficient} on the linear term but does not change the scaling class. The reason is that even with perfect verifiability, the specification cost $H_{\text{spec}} = \nu \cdot E$ remains: knowing whether an output is correct does not help you know what the correct output \emph{is}. Verification reduces $c_c$ and $c_v$ but not $\nu$.

\subsection{The March of Nines}

Karpathy's ``march of nines'' \citep{karpathy2025verify} provides a complementary mechanism for why $H$ remains linear. The phrase refers to the observation that each additional ``nine'' of per-step reliability (from 90\% to 99\% to 99.9\%, and so on) requires roughly equal engineering effort to achieve, yet the compounding effect of unreliability across steps is devastating. In a multi-step workflow of $E$ steps, if each step succeeds independently with probability $p$, end-to-end reliability is $p^E$. At $p = 0.99$ and $E = 100$, end-to-end success is only $36.6\%$. At $p = 0.95$, it drops to $0.6\%$.

To maintain end-to-end reliability above a target $R$, the human must checkpoint every $k = \lfloor \log R / \log p \rfloor$ steps. The number of checkpoints is $E/k$---linear in $E$ for any fixed $p$ and $R$. Moreover, each additional ``nine'' of per-step reliability (from 90\% to 99\% to 99.9\%) requires roughly equal engineering effort, making it expensive to increase $k$. This means that even as agents improve, the checkpoint frequency decreases only logarithmically with capability while the task length $E$ grows linearly, keeping total checkpoint effort $O(E)$.

\subsection{Scope: Cognitive Effort vs.\ Physical Action}
\label{sec:scope}

Our model treats human effort as entirely \emph{cognitive}: specifying, verifying, correcting, and decomposing decisions. This is a reasonable approximation for knowledge work like software engineering, where ``doing'' is mostly ``deciding'' and the agent executes decisions near-instantaneously.

But many domains involve irreducible physical or temporal actions that AI agents cannot shortcut regardless of their cognitive capability: running wet-lab experiments, waiting for clinical trial results, deploying hardware, shipping products, conducting user interviews, or navigating regulatory approval. These actions introduce a separate bottleneck that our model does not capture.

This omission makes our analysis \emph{conservative}. The cognitive bottleneck we model is a lower bound on total human effort; the physical/temporal bottleneck is additive. In domains like drug discovery, the physical bottleneck may dominate: even if an AI agent could design the perfect drug candidate in seconds (eliminating all cognitive novelty), the clinical trials still take years. Amodei acknowledges this in ``Machines of Loving Grace,'' noting that ``the physical world'' and ``human societal factors'' create irreducible delays beyond what intelligence can compress.

For the remainder of the paper, $H$ refers to cognitive human effort. In domains where physical action is a significant component, total human involvement will be higher than our model predicts.

\subsection{Relationship to Amdahl's and Brooks's Laws}
\label{sec:amdahl}

Our model is structurally identical to Amdahl's Law \citep{amdahl1967}: given a serial fraction $(1-p)$, the maximum speedup from $N$ processors is $S(N) = 1/((1-p) + p/N)$, bounded by $1/(1-p)$ even as $N \to \infty$. In our framework, ``processors'' are AI agents, ``parallelizable work'' is routine subtasks, and the novelty fraction $\nu$ plays the role of $(1-p)$. It also connects to Brooks's Law \citep{brooks1975}: adding people to a late project increases communication overhead quadratically. In our setting, adding agents to a novel task doesn't help because the bottleneck is serial human cognition, not parallelizable labor. Gustafson's Law \citep{gustafson1988} offers the counterpoint: use agents not to solve fixed problems faster, but to tackle \emph{more} routine problems in the same time---exploitation scaling, not exploration scaling.

\section{Simulation Results}
\label{sec:simulations}

The simulations below do not ``prove'' the model---they are Monte Carlo instantiations of it. Their purpose is twofold: (1) to verify that the derived consequences (no sublinear regime, coefficient-not-exponent improvement) actually follow from the assumptions under realistic parameterizations, and (2) to provide quantitative estimates of the $H/E$ ratio across configurations. All code is available at \url{https://github.com/jacky-liang/novelty-bottleneck}.

\subsection{Simulation Design}
\label{sec:sim-design}

\subsubsection*{Per-decision simulation procedure}

We simulate a task of effort $E$ as a sequence of $E$ independent decision points. At each decision point $i$:

\begin{enumerate}
\item \textbf{Classify}: Draw $u_i \sim \text{Uniform}(0,1)$. If $u_i < \nu$, classify decision $i$ as \emph{novel}; otherwise \emph{routine}.
\item \textbf{Specify}: If novel, add specification cost: $H_{\text{spec}} \mathrel{+}= s$, where $s$ is the per-decision specification cost (default $s = 1.0$).
\item \textbf{Execute}: The agent attempts the decision. Draw $u'_i \sim \text{Uniform}(0,1)$. If novel, the agent succeeds with probability $p_{\text{novel}}$; if routine, with probability $p_{\text{routine}}$.
\item \textbf{Self-correct}: If the agent fails, draw $u''_i \sim \text{Uniform}(0,1)$. If $u''_i < r$ (self-correction rate), the error is caught internally. Otherwise, the error accumulates.
\end{enumerate}

After all $E$ decisions, human effort is computed as:
$H = H_{\text{spec}} + c_v \cdot E + c_c \cdot n_{\text{errors}},$
where $c_v$ is the per-decision verification cost and $c_c$ is the per-error correction cost.

\subsubsection*{Experiment configurations}

Table~\ref{tab:configs} specifies the four agent configurations used in the scaling experiment. All configurations share $c_c = 2.0$ (correction cost) and $s = 1.0$ (specification cost).

\begin{table}[h]
\centering
\caption{Agent configuration parameters for the scaling experiment.}
\begin{tabular}{lccccc}
\toprule
Configuration & $\nu$ & $p_{\text{routine}}$ & $p_{\text{novel}}$ & $r$ & $c_v$ \\
\midrule
Low Novelty & 0.1 & 0.95 & 0.3 & 0.0 & 0.05 \\
Medium Novelty & 0.3 & 0.95 & 0.3 & 0.0 & 0.05 \\
High Novelty & 0.8 & 0.95 & 0.3 & 0.0 & 0.05 \\
High Capability & 0.3 & 0.99 & 0.7 & 0.8 & 0.02 \\
\bottomrule
\end{tabular}
\label{tab:configs}
\end{table}

\subsubsection*{Experimental protocol}

For each configuration, we simulate tasks at $E \in \{10, 25, 50, 100, 200, 500, 1000, 2000, 5000\}$, with 50 independent trials per $(E, \text{config})$ pair. All experiments use NumPy's \texttt{default\_rng} with seed 42 for reproducibility. We report means across trials; standard deviations are small relative to means for $E \geq 100$.

\textbf{Scaling exponents} are estimated by ordinary least squares regression of $\log H$ on $\log E$ across the 9 task sizes, using trial-averaged $H$ values.

\textbf{Mutual information experiment}: We vary $M \in \{0.0, 0.3, 0.6, 0.9, 0.99\}$. At each decision, the agent infers the correct specification with probability $M$; with probability $1-M$, the human must provide it at cost $s = 1.0$. A per-decision verification cost of $c_v = 0.05$ is added regardless. Seed: 456.

\textbf{Trajectory divergence}: We simulate a random walk with step noise $\epsilon_i \sim \mathcal{N}(0, \sigma^2)$, $\sigma = 0.1$, for $E$ steps. We record the maximum absolute cumulative deviation $\max_{t \leq E} |\sum_{i=1}^t \epsilon_i|$ across 100 independent trials per $E$ value. Seed: 123.

\textbf{Novelty dominance}: This uses an analytical (non-stochastic) model. For each $\nu \in \{0, 0.01, 0.05, 0.1, 0.2, 0.5, 1.0\}$ and $E \in \{10, 20, \ldots, 2000\}$, we compute $H = \nu \cdot E + (1-\nu) \cdot 2\log_2 E + 0.05 E$, where the first term represents novel decisions, the second represents sublinear gains on routine decisions, and the third represents irreducible verification.

\textbf{March of nines}: Uses the analytical formula $R_{\text{e2e}} = p^E$ for end-to-end reliability, with checkpoint interval $k = \lfloor \log(0.8) / \log(p) \rfloor$ (the human checkpoints whenever segment reliability drops below 80\%).

\textbf{Verifiability frontier}: Computes $H/E = \nu(1 - 0.8v) + 0.05(1-v) + 0.02$, where $0.8$ is the self-correction rate when automated verification is available, $0.05(1-v)$ is the human verification cost, and $0.02$ is a base coordination cost.

Simulation parameters for the organizational scaling model (Section~\ref{sec:org}) are presented alongside that model.

\subsection{Scaling Exponents}

Table~\ref{tab:exponents} reports the fitted scaling exponents for each configuration.

\begin{table}[h]
\centering
\caption{Fitted scaling exponents $\alpha$ where $H \sim E^\alpha$.}
\begin{tabular}{lcc}
\toprule
Configuration & $\alpha$ & $H/E$ at $E{=}5000$ \\
\midrule
Low Novelty ($\nu = 0.1$) & 0.999 & 0.377 \\
Medium Novelty ($\nu = 0.3$) & 1.004 & 0.841 \\
High Novelty ($\nu = 0.8$) & 1.002 & 1.990 \\
High Capability + Self-Correction & 1.004 & 0.360 \\
\bottomrule
\end{tabular}
\label{tab:exponents}
\end{table}

The scaling exponent is $\alpha \approx 1.0$ in all cases. Improved agent capability (higher accuracy, self-correction) reduces the coefficient $c_H$ but does not change the exponent. This confirms our theoretical prediction: better agents improve constants, not asymptotics.

\subsection{Mutual Information}

Table~\ref{tab:mi} shows how the $H/E$ ratio varies with the mutual information $M$ between agent prior and human intent.

\begin{table}[h]
\centering
\caption{Effect of mutual information on $H/E$ ratio at $E = 5{,}000$.}
\begin{tabular}{cc}
\toprule
Mutual Information & $H/E$ \\
\midrule
0.0 & 1.050 \\
0.3 & 0.751 \\
0.6 & 0.450 \\
0.9 & 0.151 \\
0.99 & 0.060 \\
\bottomrule
\end{tabular}
\label{tab:mi}
\end{table}

High mutual information dramatically reduces the coefficient but the relationship $H \sim c \cdot E$ holds throughout. Even at $M = 0.99$ (agent infers 99\% of intent), $H$ is still linear in $E$. The coefficient $c$ approaches $c_v$ (pure verification cost) as $M \to 1$.

\subsection{Novelty Dominance}

\begin{figure}[h]
\centering
\includegraphics[width=0.95\textwidth]{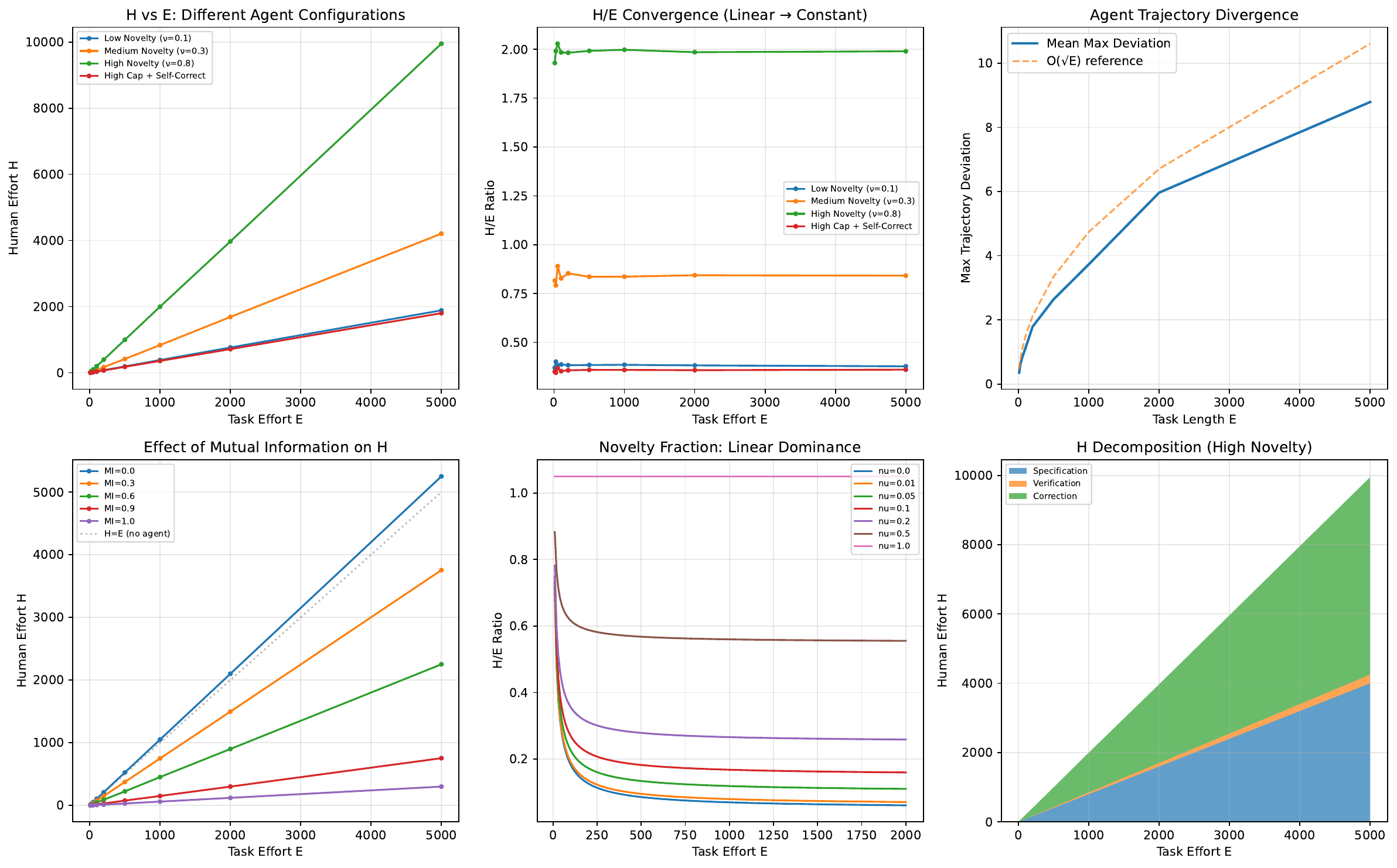}
\caption{Simulation results across configurations. \textbf{Top left}: $H$ vs.\ $E$ is linear for all configurations. \textbf{Top center}: $H/E$ ratio converges to a constant. \textbf{Top right}: Agent trajectory divergence grows as $O(\sqrt{E})$, confirming the random walk model. \textbf{Bottom left}: Higher mutual information reduces the linear coefficient but not the scaling class. \textbf{Bottom center}: Even 1\% novelty makes $H/E$ converge to a positive constant. \textbf{Bottom right}: Stacked decomposition of $H$ for the high-novelty case.}
\label{fig:novelty}
\end{figure}

Our key theoretical prediction is that any nonzero novelty fraction causes the linear term to dominate at large $E$. We modeled $H$ as a mixture: $H = \nu \cdot E + (1-\nu) \cdot \log_2(E) \cdot 2 + 0.05 E$. Figure~\ref{fig:novelty} shows that even 1\% novelty ($\nu = 0.01$) produces a visibly linear $H/E$ ratio that converges to a constant, while the $\nu = 0$ case shows the characteristic $\log(E)/E \to 0$ decay of sublinear scaling.

\begin{figure}[h]
\centering
\includegraphics[width=0.85\textwidth]{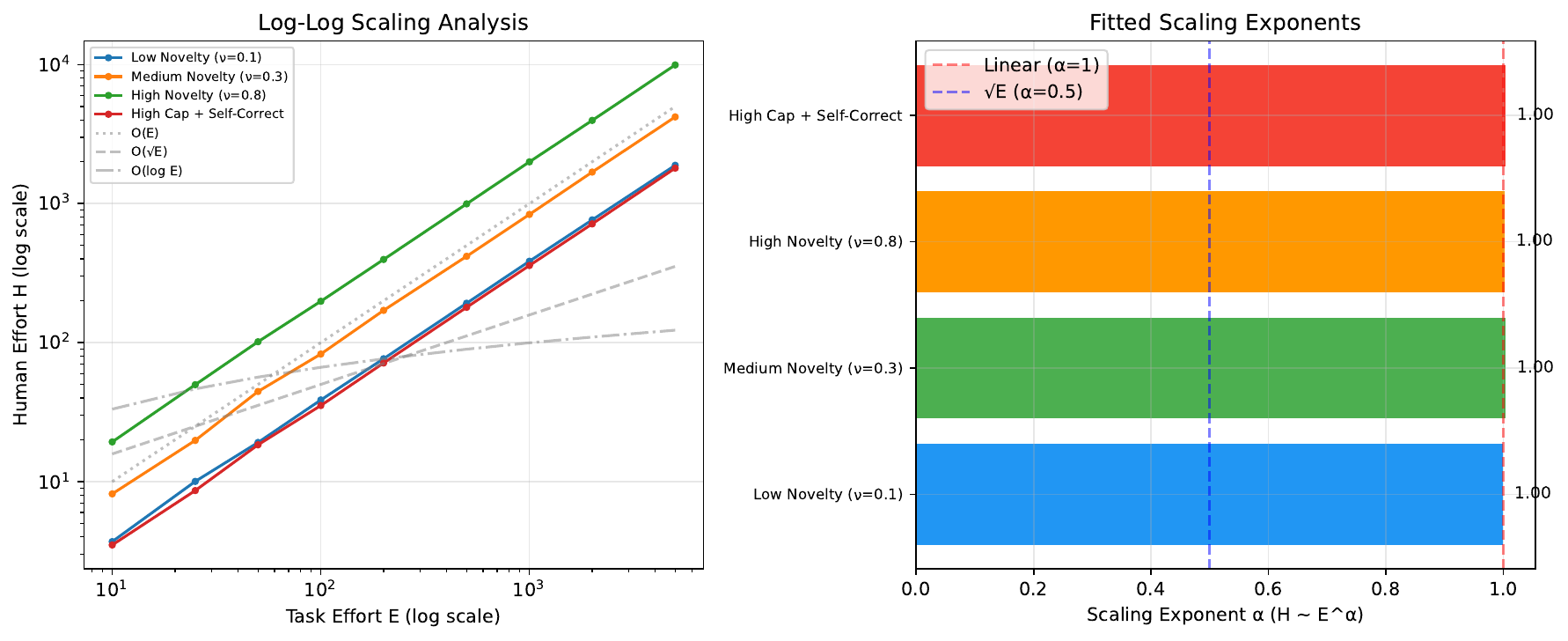}
\caption{\textbf{Left}: Log-log plot confirming linear scaling ($\alpha \approx 1$) across all configurations. Reference lines for $O(E)$, $O(\sqrt{E})$, and $O(\log E)$ shown. \textbf{Right}: Fitted scaling exponents; all configurations cluster tightly around $\alpha = 1.0$.}
\label{fig:exponents}
\end{figure}

\subsection{Reliability Compounding (March of Nines)}

\begin{figure}[ht]
\centering
\includegraphics[width=0.95\textwidth]{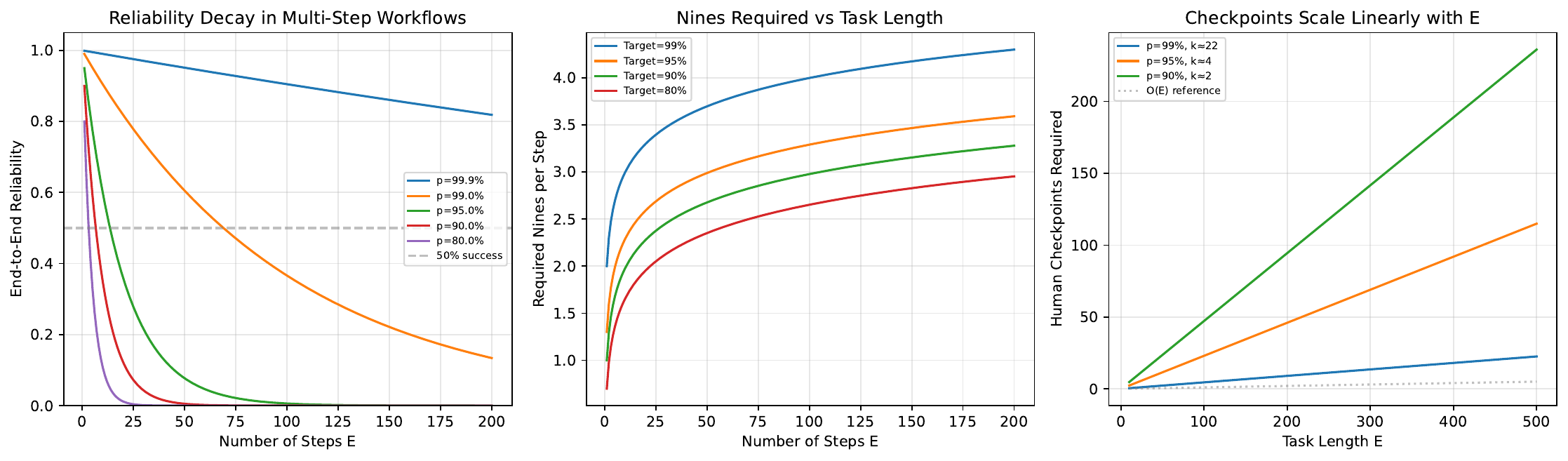}
\caption{March of nines analysis. \textbf{Left}: End-to-end reliability decays exponentially with task length. \textbf{Center}: Required per-step nines grow with $E$ (log scale). \textbf{Right}: Human checkpoints required scale linearly with $E$.}
\label{fig:nines}
\end{figure}

Figure~\ref{fig:nines} confirms the checkpoint analysis from Section~\ref{sec:model}. At per-step reliability $p = 0.95$, a 100-step workflow has end-to-end success of only $0.6\%$, requiring a human checkpoint every $\sim$4 steps. Even at $p = 0.99$, end-to-end success for 100 steps is just $36.6\%$. The number of required checkpoints scales linearly with $E$ for any fixed per-step reliability and target threshold.

\subsection{Verifiability Frontier}

\begin{figure}[ht]
\centering
\includegraphics[width=0.95\textwidth]{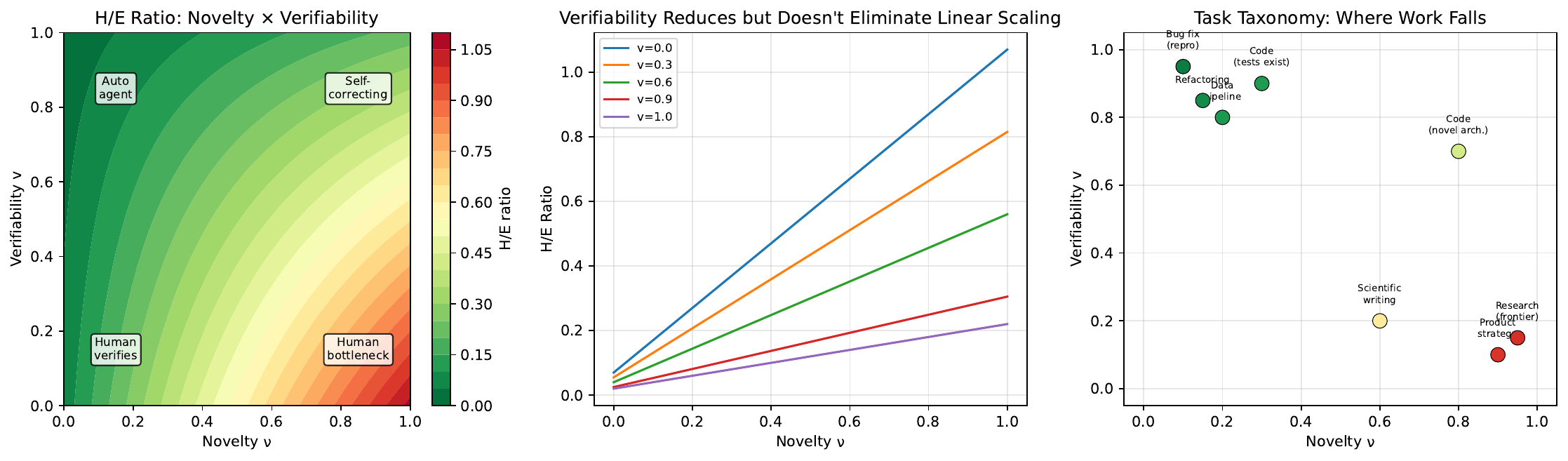}
\caption{Verifiability frontier. \textbf{Left}: $H/E$ ratio across the novelty-verifiability plane. \textbf{Center}: Verifiability reduces the slope but does not eliminate linear scaling. \textbf{Right}: Concrete task types placed in the novelty-verifiability space.}
\label{fig:verify}
\end{figure}

Figure~\ref{fig:verify} maps the two-dimensional landscape of novelty and verifiability. The heatmap shows that the lowest-effort tasks cluster in the low-novelty, high-verifiability corner (routine code with test suites), while the highest-effort tasks fall in the high-novelty, low-verifiability corner (frontier research, product strategy). Importantly, high verifiability reduces the coefficient but all quadrants with $\nu > 0$ remain linear in $E$.

\subsection{Empirical Consistency}

Several independent studies produce findings consistent with our predictions.

\textbf{The METR RCT (2025).} In a randomized controlled trial \citep{metr2025}, 16 experienced developers completed 246 tasks on their own repositories (averaging 5 years experience, 1M+ lines of code), randomly assigned to use or not use AI tools. AI made them 19\% slower (CI: $[+2\%, +39\%]$), yet developers believed they were 20\% faster. Our model explains this: these tasks were novel \emph{relative to the agent's prior} ($\nu$ was high), so specification and verification costs exceeded execution savings, making $c_H > 1$. METR's 2026 follow-up \citep{metr2026} found the experiment had become unreliable as developers refused to work without AI---suggesting AI's value is real but $\nu$-dependent.

\textbf{DORA (2024--2025).} Google's survey of $\sim$39,000 professionals found that 75\% reported feeling more productive with AI, yet every 25\% increase in AI adoption was associated with 1.5\% less delivery throughput and 7.2\% less stability \citep{dora2024}. The 2025 report confirmed persistent instability \citep{dora2025}. This maps to our march-of-nines analysis: AI increases output rate without increasing per-step reliability.

\textbf{Faros AI (2025).} Telemetry from 10,000+ developers across 1,255 teams: developers write more code, but companies see no measurable improvement in delivery velocity \citep{faros2025}---the organizational manifestation of individual $c_H$ decreasing while system-level coordination absorbs the gains.

\textbf{SWE-bench scaling.} Agents resolve 65--75\% of SWE-bench Verified tasks (routine, $<$1 hour) but only 19--21\% of SWE-EVO tasks (novel, multi-file) \citep{jimenez2024}---a sharp novelty-dependent performance drop. \textbf{Scientific productivity:} Collison and Nielsen \citep{collison2018} document that breakthroughs per dollar remain roughly constant despite exponentially more researchers. \textbf{Demo-production gap:} Karpathy's MenuGen experience \citep{karpathy2025verify}---80\% done at demo stage, 20\% done in reality---illustrates how routine progress masks the dominance of novel subtasks at scale.

\subsubsection*{Temporal limitations}

These studies reflect early-to-mid 2025 tools. Coding agents have since improved substantially: SWE-bench scores rose from 49\% to 75\%+; Claude Code matured into a production platform. Our framework predicts this should reduce $c_H$ (coefficient improvement) without changing the scaling exponent: more tasks become routine, but genuinely novel tasks remain linear. Whether this prediction holds is the most important near-term test of our model.

\section{Organizational Scaling: Multiple Humans with AI Agents}
\label{sec:org}

Our analysis so far considers a single human working with AI agents. In practice, organizations deploy $n$ humans, each paired with agents, collaborating on a shared task. This introduces a second bottleneck: human-human coordination, which interacts with the human-agent bottleneck in a way that has significant implications for how AI adoption affects organizational structure.

\subsection{Model}

Consider $n$ humans collaborating on a task of effort $E$, each paired with AI agents. Total human effort now includes three terms:
\begin{equation}
H_{\text{total}}(n) = c \cdot E + \frac{\beta \cdot n(n-1)}{2} + \frac{\gamma \cdot n(n-1)}{2} \cdot \tau,
\end{equation}
where $c \cdot E$ is the irreducible work (specification, verification, correction from Section 2), $\beta$ captures pairwise coordination cost (maintaining consistency of intent across humans --- this is Brooks's $O(n^2)$ overhead), and $\gamma \cdot \tau$ captures integration cost amplified by agent throughput $\tau$. The throughput amplifier $\tau$ reflects the fact that each human-agent pair produces output at rate $\tau$ times the rate of a human alone, and all this output must be integrated.

The wall-clock time, which organizations typically optimize for, is:
\begin{equation}
\label{eq:wallclock}
T(n) = \frac{c \cdot E}{n} + \beta \cdot n + \gamma \cdot n \cdot \tau.
\end{equation}
The first term shrinks with team size (parallelism benefit); the second and third grow (coordination and integration cost).

\subsection{Optimal Team Size}

Minimizing Eq.~\ref{eq:wallclock} with respect to $n$ yields:
\begin{equation}
\label{eq:optimal_n}
n^* = \sqrt{\frac{c \cdot E}{\beta + \gamma \cdot \tau}},
\end{equation}
with corresponding minimum wall-clock time:
\begin{equation}
T^* = 2\sqrt{c \cdot E \cdot (\beta + \gamma \cdot \tau)} \sim O(\sqrt{E}).
\end{equation}

We analyze four configurations representing increasing agent capability. All are computed analytically (no stochastic simulation).

\begin{table}[h]
\centering
\caption{Organizational scaling configuration parameters.}
\begin{tabular}{lcccc}
\toprule
Configuration & $c$ & $\beta$ & $\gamma$ & $\tau$ \\
\midrule
No AI & 1.0 & 0.5 & 0.0 & 1.0 \\
Basic AI ($2\times$) & 0.5 & 0.5 & 0.1 & 2.0 \\
Strong AI ($5\times$) & 0.2 & 0.5 & 0.1 & 5.0 \\
Frontier AI ($10\times$) & 0.1 & 0.5 & 0.1 & 10.0 \\
\bottomrule
\end{tabular}
\label{tab:org_configs}
\end{table}

This gives two key results.

\textbf{Result 1: Better agents shrink optimal team size.} As agent throughput $\tau$ increases, $n^*$ decreases as $O(1/\sqrt{\tau})$. More capable agents make each human more productive, reducing the need for additional humans, while the coordination cost of each additional human grows (because they produce more output that must be integrated). Table~\ref{tab:org} confirms this: optimal team size drops from 100 to 18 as agent capability increases from no AI to frontier AI.

\begin{table}[ht]
\centering
\caption{Optimal team size and minimum wall-clock time for $E = 5{,}000$.}
\begin{tabular}{lccc}
\toprule
Configuration & Throughput $\tau$ & $n^*$ & $T^*$ \\
\midrule
No AI & $1\times$ & 100.0 & 100.0 \\
Basic AI & $2\times$ & 59.8 & 83.7 \\
Strong AI & $5\times$ & 31.6 & 63.2 \\
Frontier AI & $10\times$ & 18.3 & 54.8 \\
\bottomrule
\end{tabular}
\label{tab:org}
\end{table}

\textbf{Result 2: Wall-clock time scales as $O(\sqrt{E})$, but total effort remains $O(E)$.} At the optimal team size, wall-clock time achieves sublinear scaling in $E$ --- a genuine parallelism benefit. But total human effort across all team members is:
\begin{equation}
H_{\text{total}}(n^*) = c \cdot E + O((n^*)^2) = c \cdot E + O(E) = O(E).
\end{equation}
The coordination overhead at optimal team size is itself $O(E)$, so total human effort is linear regardless. You can trade wall-clock time for total effort, but you cannot escape the linear total.

\begin{figure}[ht]
\centering
\includegraphics[width=0.95\textwidth]{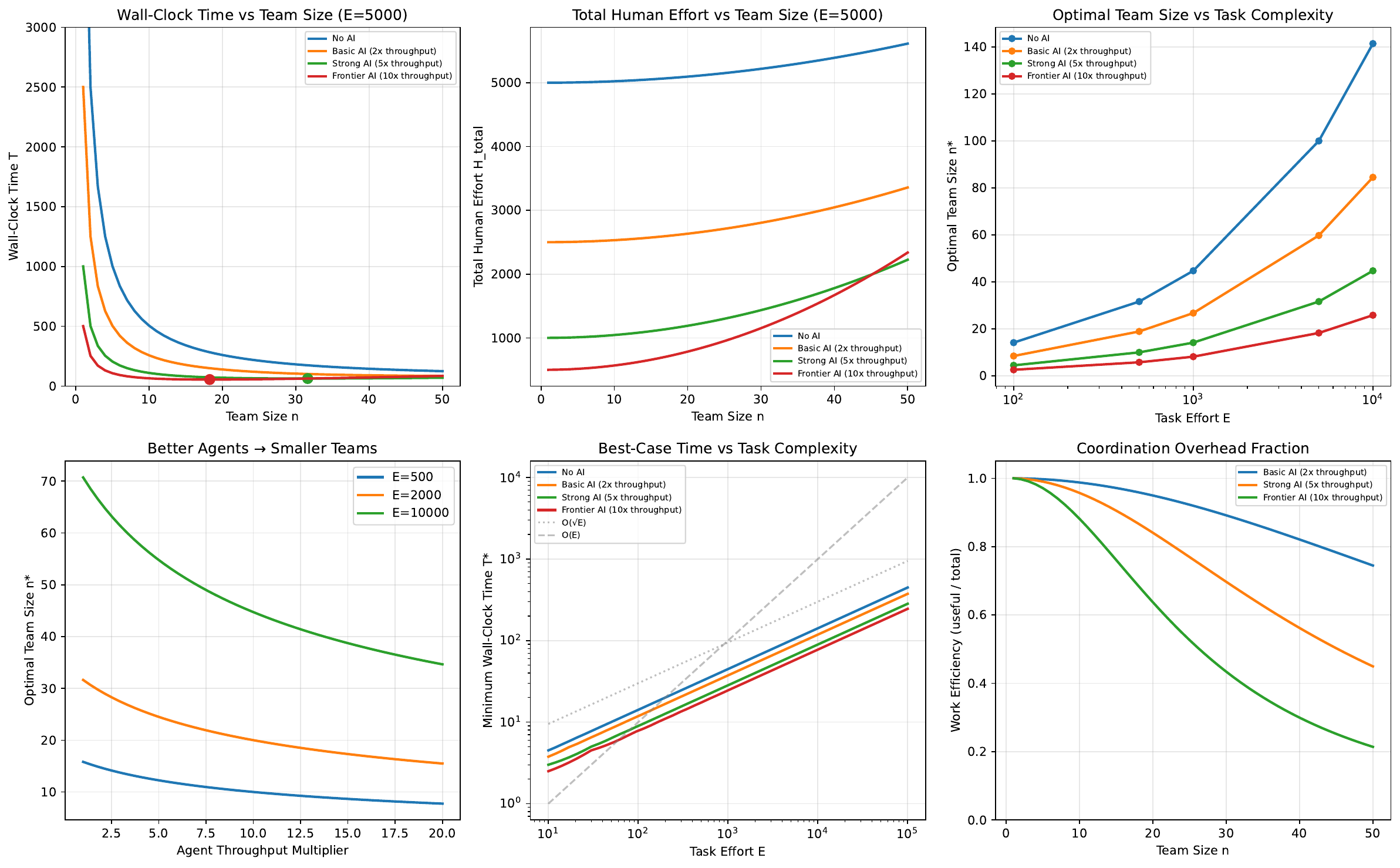}
\caption{Organizational scaling results. \textbf{Top left}: Wall-clock time vs.\ team size shows U-shaped curves with optimal $n^*$ (dots) that shifts left as agents improve. \textbf{Top center}: Total human effort grows superlinearly past $n^*$. \textbf{Top right}: Optimal team size grows as $O(\sqrt{E})$ but shrinks with agent capability. \textbf{Bottom left}: Better agents systematically reduce optimal team size. \textbf{Bottom center}: Minimum achievable wall-clock time scales as $O(\sqrt{E})$ for all configurations. \textbf{Bottom right}: Work efficiency (fraction of effort spent on useful work vs.\ coordination) decays rapidly with team size, faster for more capable agents.}
\label{fig:org}
\end{figure}

\subsection{The AI-Era Brooks's Law}

This yields what we call the \emph{AI-era Brooks's Law}: \textbf{AI agents make large teams less efficient, not more.} In the pre-AI regime, adding a person adds throughput $1$ and coordination cost $O(n)$. In the AI regime, each person commands agents that amplify their throughput to $\tau$, but this amplified output increases the integration burden on every other team member. The marginal coordination cost of person $n+1$ is $\beta + \gamma \cdot \tau$, which grows with agent capability.

The practical prediction is that AI-leveraged organizations should converge toward \emph{smaller, more autonomous teams} rather than larger ones. A team of 5 engineers each using frontier AI agents can match the output of a team of 50 without agents, while incurring far less coordination overhead. This is consistent with emerging empirical observations of small AI-augmented startups outperforming larger traditional engineering organizations.

The deeper structural point is that the human-human coordination bottleneck and the human-agent specification bottleneck are \emph{multiplicative}: agent amplification makes the coordination bottleneck worse, not better. AI does not solve the management problem; it intensifies it.

\section{Critical Examination and Conditions for Falsification}
\label{sec:critique}

We stress-test our claims by considering the strongest counterarguments. Our thesis describes the \emph{current and near-term regime} where AI agents execute human intent. We claim $H \sim O(E)$ holds while (a) agents don't learn mid-task, (b) humans are the source of intent, and (c) verification of non-verifiable outputs requires human judgment. Each counterargument below targets one of these assumptions.

\subsection{The Continual Learning Escape}

If agents could learn during task execution---updating their prior from each interaction, as humans do---then effective novelty $\nu(t)$ would decay, potentially yielding $H_{\text{spec}} \sim O(\log E)$. Ilya Sutskever, co-founder of Safe Superintelligence Inc., argues that current models ``generalize dramatically worse than people'' and that this is a solvable research problem \citep{sutskever2025}. \textbf{Condition for falsification}: $\nu(t) \sim t^{-\alpha}$ for $\alpha > 0$ within a single task. \textbf{Assessment}: No current system demonstrates this; we treat it as the most likely route to eventual falsification, requiring an architectural breakthrough rather than incremental scaling.

\subsection{Other Counterarguments}

\textbf{Autonomous exploration.} If agents can explore, hypothesize, and self-verify without human input, $H$ could decouple from $E$. But autonomous exploration requires verification of novel results---either a trusted oracle or human judgment---and uncomprehended output is noise, not work. This escape is plausible in formally verifiable domains (mathematics, certain engineering) but unlikely where taste, judgment, or real-world grounding matter.

\textbf{Cognitive primitives.} Nielsen \citep{nielsen2024} argues AI might create new cognitive operations that restructure what counts as routine---a phase transition, not linear speedup. This is compatible with our model if the transition operates by reducing $\nu$ discontinuously. Historical cognitive tools (writing, algebra, programming) took decades to diffuse; we cannot rule out faster transitions but note the precedent.

\textbf{Amortization across tasks.} Over many similar tasks, shared structure reduces the novel fraction. This is the primary mechanism by which agents improve over time, but it operates by moving the novelty boundary, not changing the scaling law. Dwarkesh Patel's ``Napoleon argument'' \citep{patel2023} reinforces this: if parallelism helped, labs would already redirect thousands of researchers to their hardest problems.

\subsection{The Optimization--Creation Spectrum}

This counterargument reveals the precise boundary of our framework. Consider: ``Given \$10k, build an app that generates \$100k ARR.'' Specification is $O(1)$; verification is $O(1)$. If the human genuinely has no intermediate preferences, $H$ really is $O(1)$. This is not hypothetical---AlphaFold, hyperparameter search, and chip design optimization are precisely this regime, and AI already dominates there.

But most goals carry \emph{latent preferences} that surface during execution. The \$100k ARR goal sounds preference-free, but: Would you accept a legally gray app? One that collapses at month 6? One targeting a market you can't support? These are constraints you don't know you have until you react to the agent's proposal. Each revealed preference is a specification bit, and their count scales with execution complexity.

This yields a spectrum:
\begin{itemize}
\item \textbf{Pure optimization} (fully specified objective, machine-verifiable): $H = O(1)$. AI dominates.
\item \textbf{Constrained optimization} (objective exists, latent constraints surface): $H$ likely $O(E)$.
\item \textbf{Creative work} (vague objective, dense preferences, human-judgment verification): $H = O(E)$.
\end{itemize}

Our claim is not that $H = O(E)$ universally, but that the pure-optimization regime is narrower than AI discourse assumes. The hierarchy compresses the \emph{description} of intent, not the \emph{quantity}.

\section{Implications}
\label{sec:implications}

\subsection{For the ``Country of Geniuses'' Vision}

Our framework does not refute Amodei's vision \citep{amodei2024}---it explains its structure. AI agents can parallelize and accelerate the \emph{routine} experimental and computational work that constitutes the majority of scientific labor: running assays, screening compounds, optimizing protocols, analyzing data. A 10$\times$ speedup on this routine fraction is plausible and would be genuinely transformative---it would represent one of the largest accelerations in the history of science. Amodei's estimate of compressing a century of biological progress into 5--10 years is consistent with this: if roughly 90\% of scientific labor is routine execution and 10\% is serial insight generation, Amdahl's Law bounds the overall speedup to $\sim$10$\times$, yielding exactly the 5--10 year timeline.

What our framework clarifies is why the estimate is 5--10 \emph{years} rather than 5--10 \emph{months}. The insight chain---the serial sequence of novel conceptual breakthroughs that connect routine phases---cannot be compressed by adding more agents. Each breakthrough depends on understanding the results of the previous one. This is not a limitation of AI capability but of the logical structure of cumulative discovery.

The organizational implication is that the ``country of geniuses'' functions most effectively as a small team of serial thinkers directing a massive execution apparatus---which is, historically, how every major scientific project has actually worked (the Manhattan Project, the Human Genome Project, modern drug discovery). AI dramatically improves the execution layer. Our organizational analysis (Section~\ref{sec:org}) formalizes this: the optimal team size \emph{shrinks} as agent capability grows, because the coordination cost of adding more thinkers exceeds the marginal benefit once each is sufficiently AI-augmented. The progress is real and large; it is the parallelism that is bounded.

\subsection{For Recursive Self-Improvement}

Aschenbrenner's intelligence explosion scenario \citep{aschenbrenner2024} requires AI systems to recursively improve themselves faster than human oversight can track. Our model predicts that genuine architectural innovation in AI --- the kind that yields qualitative capability jumps --- is precisely the type of novel work where $\nu \approx 1$ and $H \sim O(E)$ (where ``$H$'' is now the serial cognitive effort of the system itself or its human overseers).

The routine aspects of ML engineering (hyperparameter tuning, scaling experiments, code optimization) can certainly be automated and parallelized. But the key question is whether the \emph{novel} aspects --- new architectures, new training paradigms, new theoretical insights --- can be generated through parallelism. Our analysis says no: insight generation is fundamentally serial, and adding more agent instances doesn't help.

This doesn't preclude rapid AI progress in calendar time. An AI system that ``thinks'' 100$\times$ faster than a human researcher could compress a year of serial insight generation into a few days. The speedup comes from clock speed, not parallelism. But this yields a ``fast'' takeoff in human time while still being ``slow'' in AI-subjective time --- a crucial distinction for safety considerations.

\subsection{For AI Safety and Existential Risk}

Our framework yields a nuanced AI safety picture. The reassuring implication is that a misaligned AI cannot rapidly bootstrap novel capabilities because novel capability development is serial. There would be warning signs.

The concerning implication is that \emph{exploiting existing knowledge} is mostly routine work where $\nu \approx 0$. Weaponizing known biology, executing known cyber attacks, and deploying known manipulation techniques are all tasks where the agent has enormous prior. The novelty bottleneck constrains frontier research, not the exploitation of existing knowledge.

This suggests that the near-term AI safety profile is not ``sudden godlike intelligence'' but rather ``terrifyingly effective at exploiting everything humanity has already discovered, while mediocre at pushing the frontier''---a profile that does not require capability breakthroughs beyond what we have already observed.

\subsection{For Work, Employment, and Business}

Our framework connects to Acemoglu and Restrepo's task-based models \citep{acemoglu2024}, where AI's GDP impact depends on the fraction of tasks affected and their cost savings. We characterize \emph{which} tasks are affected: those where $\nu$ is low and $v$ is high.

\textbf{Task reallocation, not job elimination.} Most jobs bundle tasks across the novelty-verifiability spectrum. AI automates the routine subtasks \emph{within} jobs (low $\nu$, high $v$), concentrating human effort on the novel core. This is consistent with the WEF Future of Jobs Report \citep{wef2025}, which projects tasks roughly evenly split between human, machine, and hybrid by 2030.

\textbf{Polarization along the novelty axis.} Workers whose jobs are primarily routine and verifiable (data entry, basic coding, translation) face displacement. Those with high-novelty components (strategy, design, research) are buffered by the linear bottleneck. But our model adds a nuance: \emph{workers who can reduce $\nu$ for others}---writing specifications, designing test suites, architecting for agent delegation---become more valuable. This is Acemoglu's ``new task creation'': AI generates demand for orchestration and verification meta-tasks that did not previously exist.

\textbf{Smaller firms, higher leverage.} If optimal team size decreases with agent capability (Section~\ref{sec:org}), AI-era firms should be smaller and more productive per employee---Coase's theory of the firm in reverse. AI reduces individual production costs but increases per-person coordination costs, pushing the efficient boundary toward smaller organizations. This echoes Coase's theory of the firm \citep{coase1937}, which predicts that firms expand until the cost of organizing an additional transaction internally equals the cost of carrying it out on the open market; AI shifts this boundary by lowering individual production costs while raising internal coordination costs.

\textbf{The ``so-so automation'' risk.} Our framework identifies when the ``so-so automation'' risk described by Acemoglu and Restrepo \citep{acemoglu2019}---technologies that displace workers without generating large productivity gains---is highest: AI deployed on marginally routine tasks ($\nu$ just below the automation threshold), yielding small cost savings while displacing labor. Firms should prioritize the deep-routine regime where gains are large and unambiguous.

\textbf{Aggregate growth: large but not explosive.} The Penn Wharton Budget Model \citep{pennwharton2025} projects AI will raise US GDP by 1.5\% by 2035; Acemoglu estimates 0.53--0.66\% TFP growth over a decade. Our framework explains why: the novelty bottleneck limits impact to the routine fraction of economic activity. The result is a sustained improvement comparable to electricity or computing, but not the discontinuous explosion that singularity narratives predict.

\section{Related Work}
\label{sec:related}

Our framework builds on three established results. \textbf{Amdahl's Law} \citep{amdahl1967} bounds speedup from parallelism by the serial fraction; our novelty parameter $\nu$ plays this role (Section~\ref{sec:amdahl}). Gustafson's Law \citep{gustafson1988} offers the counterpoint of scaling problem size rather than shrinking execution time. \textbf{Brooks's Law} \citep{brooks1975} shows communication overhead scales quadratically with team size; our organizational model extends this to human-AI teams. Brooks's ``No Silver Bullet'' \citep{brooks1986} distinguishes essential from accidental complexity; AI removes accidental complexity, but essential complexity---our ``novelty''---remains.

On the empirical side, \textbf{Acemoglu's} task-based model \citep{acemoglu2024} estimates 4.6\% of tasks are profitably automatable, yielding modest TFP gains---consistent with our framework's prediction that impact is bounded by the routine fraction. \textbf{Karpathy's} verifiability frontier and march of nines \citep{karpathy2025verify} provide practitioner grounding. \textbf{Collison and Nielsen} \citep{collison2018} document declining research productivity per dollar; Nielsen's discovery fiction \citep{nielsen2024} provides epistemological foundations for our serial-insight claim. \textbf{SWE-bench} \citep{jimenez2024} and its successors show sharp performance drops as task novelty increases. \textbf{Scaling laws} for language models \citep{kaplan2020} describe capability growth; our work addresses how \emph{deploying} capabilities on novel tasks faces structural bottlenecks that scaling alone cannot overcome. For AI progress forecasts, we engage with Aschenbrenner \citep{aschenbrenner2024}, Amodei \citep{amodei2024}, and Patel \citep{patel2023} throughout.

\section{Discussion}
\label{sec:discussion}

The central finding of this paper is a negative result: under current architectures, there is no intermediate regime between ``human does everything'' and ``human does nothing'' where human effort scales sublinearly with task complexity. For any task containing genuine novelty, $H \sim O(E)$.

We want to be precise about what this does and does not imply.

\textbf{What it does not imply.} It does not imply that AI agents are not transformative. Reducing the constant $c_H$ from 1.0 (no agent) to 0.1 (excellent agent) is a 10$\times$ productivity gain---enormous by any historical standard. Karpathy has estimated that AI-assisted development will keep GDP growth at $\sim$2\% rather than triggering a discontinuous acceleration \citep{karpathy2025verify}; this is consistent with our model: a large constant-factor improvement spread across the economy, not a change in growth regime. It also does not imply that the bottleneck is permanent. Section~\ref{sec:critique} identifies three specific conditions (continual learning, autonomous verification, new cognitive primitives) under which the scaling class could change. Our claim is that these require architectural breakthroughs, not incremental scaling of current approaches.

\textbf{What it does imply.} The organizational analysis in Section~\ref{sec:org} shows that even team parallelism cannot escape the linear total: wall-clock time achieves $O(\sqrt{E})$ but total human effort remains $O(E)$, and better agents push organizations toward smaller teams, not larger ones. The ``country of geniuses'' vision is achievable---but it achieves its gains through constant-factor compression of the routine fraction, not through superlinear scaling. Understanding this distinction matters for setting expectations and designing organizational structures.

\textbf{The dynamic boundary.} Perhaps our most important concession is that the novelty fraction $\nu$ is not fixed. It shrinks over time as agent priors expand, as new tools restructure cognition (Nielsen's cognitive primitives), and as formerly novel tasks become routine through accumulated experience. This compounding reclassification is the primary source of AI's long-run value. Our model says $H \sim \nu E$ where $\nu$ itself is a decreasing function of cumulative agent-human interaction. The question is whether $\nu$ decreases smoothly (gradual improvement, consistent with our thesis) or undergoes a phase transition (sudden capability jump, potentially falsifying it). Historical precedent with cognitive tools---writing, calculus, programming---suggests smooth decrease over decades, but we cannot rule out discontinuity.

The deepest version of our argument is this: for truly novel work, the human doesn't know what they want until they see partial results. Intent is not a fixed input to the process; it is \emph{generated by} the process. This means human effort is not specification-as-input but specification-as-reaction, and it cannot be compressed below the rate at which novel information is produced. This is the irreducible serial core of creative and intellectual work.

\subsection{Practical Recommendations}

Our framework suggests concrete strategies. The core insight: \emph{maximize the routine, verifiable fraction and keep the novel component small.}

\textbf{1. Reduce $\nu$ systematically.} Convert novel decisions into routine ones \emph{before} delegating to agents: invest in conventions, test suites, style guides, and architectural patterns that encode intent. Systematically capture what works across tasks---reusable prompts, agent configurations, shared context---so that $\nu$ shrinks over time. Organizations that treat reducing $\nu$ as a first-class engineering objective will compound their AI leverage.

\textbf{2. Maximize verifiability.} Write tests before delegating. Define acceptance criteria. Use type systems, linters, and CI pipelines as automated verification. The DORA finding that AI adoption decreased stability by 7.2\% \citep{dora2024} likely reflects deploying agents without proportionally increasing verification infrastructure. Use the novelty-verifiability quadrant (Figure~\ref{fig:verify}) to triage: delegate aggressively where verifiable; maintain human oversight where not.

\textbf{3. Prefer small teams.} The organizational result implies shrinking teams, not growing them. A team of 3--5 with frontier agents outperforms 20 with weaker tooling, because coordination overhead dominates marginal contribution at scale.

\textbf{4. Decompose along the novelty boundary.} Separate each project into a small novel core (architecture, design) handled by the human, and a large routine periphery (implementation, testing, documentation) delegated to agents. The human's role shifts from implementer to architect.

Whether this remains true for AI systems that develop their own intent --- systems that are not executing human goals but pursuing their own --- is a question that goes beyond our framework. But for the foreseeable regime where AI agents are tools deployed on human objectives, our model suggests the right question is not ``will AI make human effort unnecessary?'' but ``how fast is $\nu$ shrinking, and for which tasks?'' The answer to that question---not the scaling exponent, which our model assumes more than derives---is what will determine whether the next decade looks like gradual productivity improvement or a discontinuous transformation.

\section*{Acknowledgments}
This paper emerged from a conversation exploring the practical experience of working with AI coding agents and the gap between that experience and popular narratives about AI-driven progress.

\paragraph{AI Contribution Statement.} This paper was developed collaboratively with Claude Opus 4.6 (Anthropic, March 2026). The AI contributed to: formalization of the mathematical model, implementation of all simulation code, literature search and synthesis, drafting and iterative revision of the manuscript, and generation of all figures. The human author (J.L.) provided the original observation and core hypothesis, directed the research agenda, shaped the arguments through iterative dialogue, critically evaluated all claims, and approved the final manuscript. All intellectual positions, errors, and editorial decisions are the responsibility of the human author. The collaborative process itself---in which the human's serial cognitive effort scaled linearly with the paper's complexity---is, we note, a data point consistent with the paper's thesis.

\bibliographystyle{plainnat}

\begin{thebibliography}{99}

\bibitem[Amdahl(1967)]{amdahl1967}
Gene M. Amdahl.
\newblock Validity of the single processor approach to achieving large scale computing capabilities.
\newblock In \emph{AFIPS Spring Joint Computer Conference}, pages 483--485, 1967.

\bibitem[Amodei(2024)]{amodei2024}
Dario Amodei.
\newblock Machines of Loving Grace: How AI Could Transform the World for the Better.
\newblock Essay, October 2024.
\newblock \url{https://darioamodei.com/machines-of-loving-grace}.

\bibitem[Aschenbrenner(2024)]{aschenbrenner2024}
Leopold Aschenbrenner.
\newblock Situational Awareness: The Decade Ahead.
\newblock Essay series, June 2024.
\newblock \url{https://situational-awareness.ai}.

\bibitem[Brooks(1975)]{brooks1975}
Frederick P. Brooks, Jr.
\newblock \emph{The Mythical Man-Month: Essays on Software Engineering}.
\newblock Addison-Wesley, 1975.

\bibitem[Brooks(1986)]{brooks1986}
Frederick P. Brooks, Jr.
\newblock No Silver Bullet---Essence and Accidents of Software Engineering.
\newblock \emph{Computer}, 20(4):10--19, 1987.

\bibitem[Gustafson(1988)]{gustafson1988}
John L. Gustafson.
\newblock Reevaluating Amdahl's Law.
\newblock \emph{Communications of the ACM}, 31(5):532--533, 1988.

\bibitem[Jimenez et al.(2024)]{jimenez2024}
Carlos E. Jimenez, John Yang, Alexander Wettig, Shunyu Yao, Kexin Pei, Ofir Press, and Karthik Narasimhan.
\newblock SWE-bench: Can Language Models Resolve Real-World GitHub Issues?
\newblock In \emph{ICLR}, 2024.

\bibitem[Kaplan et al.(2020)]{kaplan2020}
Jared Kaplan, Sam McCandlish, Tom Henighan, Tom B. Brown, Benjamin Chess, Rewon Child, Scott Gray, Alec Radford, Jeffrey Wu, and Dario Amodei.
\newblock Scaling Laws for Neural Language Models.
\newblock \emph{arXiv preprint arXiv:2001.08361}, 2020.

\bibitem[Patel(2023)]{patel2023}
Dwarkesh Patel.
\newblock Will Scaling Work?
\newblock Blog post, December 2023.
\newblock \url{https://www.dwarkesh.com/p/will-scaling-work}.

\bibitem[Acemoglu(2024)]{acemoglu2024}
Daron Acemoglu.
\newblock The Simple Macroeconomics of AI.
\newblock NBER Working Paper 32487, 2024.

\bibitem[Collison and Nielsen(2018)]{collison2018}
Patrick Collison and Michael Nielsen.
\newblock Science is Getting Less Bang for Its Buck.
\newblock \emph{The Atlantic}, November 2018.

\bibitem[Karpathy(2025)]{karpathy2025verify}
Andrej Karpathy.
\newblock Verifiability.
\newblock Blog post, November 2025.
\newblock \url{https://karpathy.bearblog.dev/verifiability/}.

\bibitem[Nielsen(2024)]{nielsen2024}
Michael Nielsen.
\newblock Discovery Fiction.
\newblock Essay, 2024.
\newblock \url{https://michaelnotebook.com/df/index.html}.

\bibitem[Becker et al.(2025)]{metr2025}
Joel Becker, Nate Rush, Tom Cunningham, and David Rein.
\newblock Measuring the Impact of Early-2025 AI on Experienced Open-Source Developer Productivity.
\newblock \emph{arXiv preprint arXiv:2507.09089}, 2025.

\bibitem[Becker et al.(2026)]{metr2026}
Joel Becker, Nate Rush, Tom Cunningham, David Rein, and Khalid Mahamud.
\newblock We are Changing our Developer Productivity Experiment Design.
\newblock METR Blog, February 2026.

\bibitem[DORA(2024)]{dora2024}
DORA Research Program.
\newblock 2024 Accelerate State of DevOps Report.
\newblock Google Cloud, October 2024.

\bibitem[DORA(2025)]{dora2025}
DORA Research Program.
\newblock 2025 DORA Report.
\newblock Google Cloud, September 2025.

\bibitem[Faros AI(2025)]{faros2025}
Faros AI.
\newblock The AI Productivity Paradox Report.
\newblock Technical report, 2025.

\bibitem[World Economic Forum(2025)]{wef2025}
World Economic Forum.
\newblock Future of Jobs Report 2025.
\newblock Geneva, January 2025.

\bibitem[Penn Wharton(2025)]{pennwharton2025}
Penn Wharton Budget Model.
\newblock The Projected Impact of Generative AI on Future Productivity Growth.
\newblock Technical report, September 2025.

\bibitem[Sutskever(2025)]{sutskever2025}
Ilya Sutskever.
\newblock Sequence to Sequence Learning: Beyond the Obvious (NeurIPS 2024 keynote).
\newblock December 2024.

\bibitem[Coase(1937)]{coase1937}
Ronald H. Coase.
\newblock The Nature of the Firm.
\newblock \emph{Economica}, 4(16):386--405, 1937.

\bibitem[Acemoglu and Restrepo(2019)]{acemoglu2019}
Daron Acemoglu and Pascual Restrepo.
\newblock Automation and New Tasks: How Technology Displaces and Reinstates Labor.
\newblock \emph{Journal of Economic Perspectives}, 33(2):3--30, 2019.

\end{thebibliography}

\end{document}